\documentclass{article}

\usepackage[nonatbib, preprint]{neurips_2023}
\usepackage{amsmath}
\usepackage{caption}
\usepackage{subcaption}
\usepackage{graphicx}
\usepackage{algorithm2e}
\usepackage{float}
\usepackage[sectionbib]{chapterbib}

\newcommand{\figref}[1]{Fig.~\ref{fig:#1}}

\newcommand{\tblref}[1]{Table~\ref{tbl:#1}}

\usepackage[utf8]{inputenc} 
\usepackage[T1]{fontenc}    
\usepackage{hyperref}       
\usepackage{url}            
\usepackage{booktabs}       
\usepackage{amsfonts}       
\usepackage{nicefrac}       
\usepackage{microtype}      
\usepackage{xcolor}         

\title{Trustworthy AI in Digital Health: A Comprehensive Review of Robustness and Explainability}

\author{%
  Abdullah Mamun$^{1,2,*}$
  \And
  Shovito Barua Soumma$^{1,2}$
  \And
  Hassan Ghasemzadeh$^1$
  \AND
  \\
$^{1}$College of Health Solutions, Arizona State University, Phoenix, AZ, USA \\
  $^{2}$School of Computing and Augmented Intelligence, Arizona State University, Tempe, AZ, USA \\
  $^{*}$Correspondence: \texttt{a.mamun@asu.edu}
}

\begin{document}

\maketitle

\begin{abstract}
Ensuring trust in AI systems is essential for the safe and ethical integration of machine learning systems into high-stakes domains such as digital health. Key dimensions, including robustness, explainability, fairness, accountability, and privacy, need to be addressed throughout the AI lifecycle, from problem formulation and data collection to model deployment and human interaction. While various contributions address different aspects of trustworthy AI, a focused synthesis on robustness and explainability, especially tailored to the healthcare context, remains limited. This review addresses that need by organizing recent advancements into an accessible framework, highlighting both technical and practical considerations. We present a structured overview of methods, challenges, and solutions, aiming to support researchers and practitioners in developing reliable and explainable AI solutions for digital health. This review article is organized into three main parts. First, we introduce the pillars of trustworthy AI and discuss the technical and ethical challenges, particularly in the context of digital health. Second, we explore application-specific trust considerations across domains such as intensive care, neonatal health, and metabolic health, highlighting how robustness and explainability support trust. Lastly, we present recent advancements in techniques aimed at improving robustness under data scarcity and distributional shifts, as well as explainable AI methods ranging from feature attribution to gradient-based interpretations and counterfactual explanations. This paper is further enriched with detailed discussions of the contributions toward robustness and explainability in digital health, the development of trustworthy AI systems in the era of LLMs, and various evaluation metrics for measuring trust and related parameters such as validity, fidelity, and diversity.

\end{abstract}

\section{Introduction}
Trustworthy AI focuses on ensuring that AI systems are reliable, transparent, and accountable. By emphasizing fairness, safety, and explainability, it fosters user confidence and promotes ethical decision-making. This approach is essential for ensuring that AI-driven solutions are both effective and equitable, ultimately contributing to better outcomes in a variety of applications across industries.

\begingroup
\renewcommand\thefootnote{}
\footnotetext{This is a preprint of the paper accepted for publication in Progress in Biomedical Engineering (Impact Factor: 7.7). The final peer-reviewed version is available at \url{https://doi.org/10.1088/2516-1091/ae4e74}}
\endgroup

\begin{figure}[b!]
    \centering
    \includegraphics[width=0.7\linewidth]{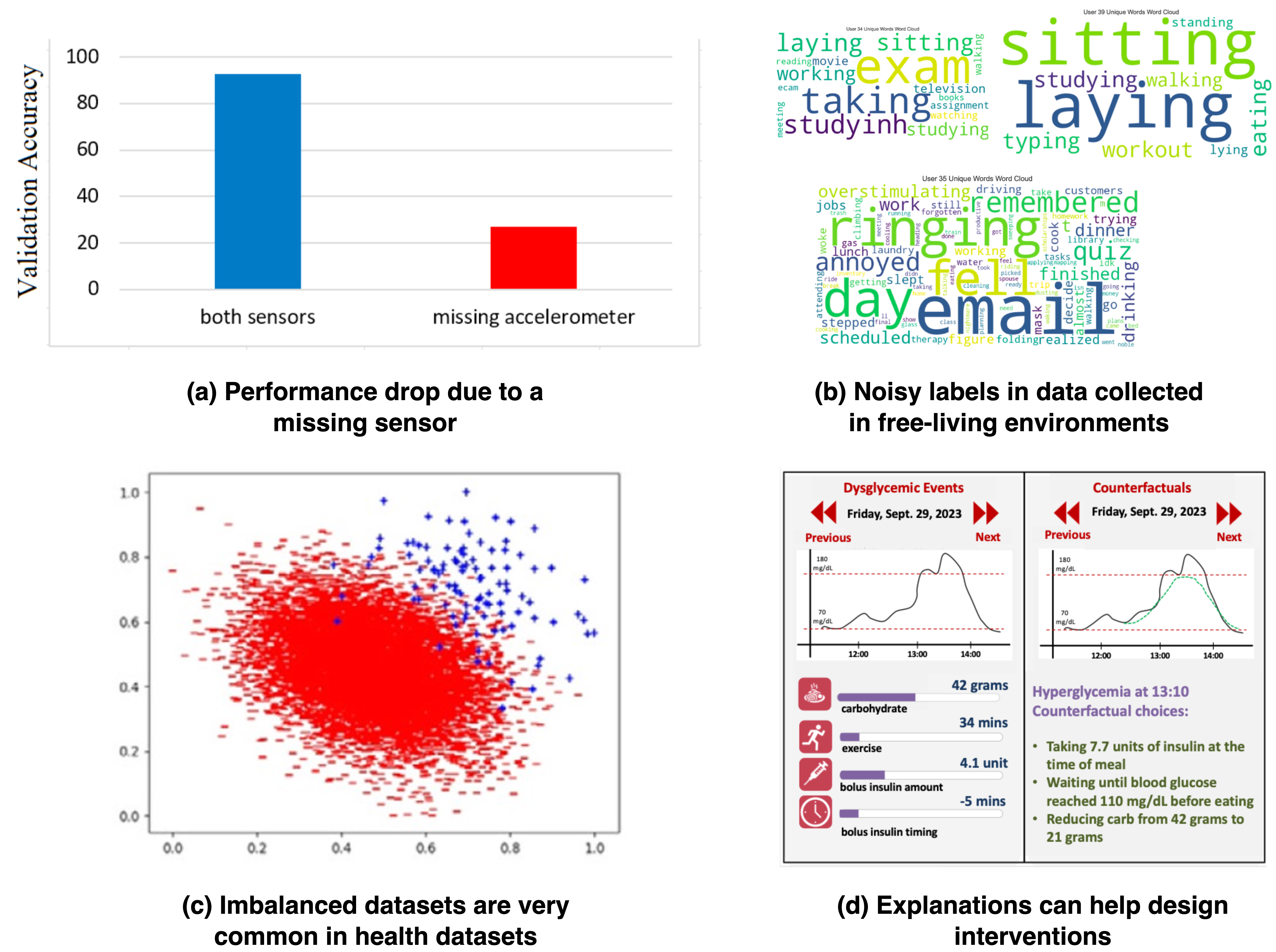}
    \caption{Examples implicating the importance of robustness and explainability in AI systems.}
    \label{fig:importance_figures}
\end{figure}

The National Institute of Standards and Technology (NIST) defines trustworthy AI through a set of core characteristics that include validity and reliability, safety, security and resiliency, accountability, transparency, explainability, privacy, and fairness \cite{nistTrustworthyResponsible}. These components serve as foundational pillars for developing standards, guidelines, and practices to ensure AI systems are aligned with ethical and societal values. The High-Level Expert Group on AI presented the Ethics Guidelines for Trustworthy Artificial Intelligence \cite{europaEthicsGuidelines} in April 2019. The guidelines define trustworthy AI as systems that are lawful, ethical, and robust, both technically and socially. They outline seven key requirements for trustworthy AI: human agency and oversight, technical robustness and safety, privacy and data governance, transparency, diversity and fairness, societal and environmental well-being, and accountability. These principles emphasize empowering human decision-making, ensuring security and reliability, respecting privacy, promoting transparency, avoiding bias, fostering inclusivity and sustainability, and implementing mechanisms for responsibility and redress.

In this article, we discuss in detail two of these components: robustness and explainability. Robustness and explainability are two important characteristics of a machine learning system. Consider a multisensor system where a human activity recognition system is trained with both accelerometer and gyroscope data. During inference, if one of the two sensors (e.g., the accelerometer) is unavailable, due to hardware limitations, being disabled to conserve power, or connectivity issues, the prediction accuracy can degrade drastically (\figref{importance_figures}a).

A challenge often faced with datasets collected in free-living environments is the presence of noisy labels. In one of our user studies at Arizona State University,  we found that different participants expressed activities in different expressions (\figref{importance_figures}b). E.g., some people used the word `typing' and some other people may write `studying' for the same activity (e.g., working on a writing assignment). Furthermore, there can be typing mistakes while recording behavioral labels, e.g., `studyinh' when intending to write `studying'. A robust machine learning system would have to overcome these challenges.

Imbalanced datasets are quite common in certain health datasets, for example, diagnosing a rather uncommon disease. When training with a highly imbalanced dataset (\figref{importance_figures}c), a machine learning model may get biased toward the majority class. To overcome the issue of imbalanced classes and small datasets, data balancing methods and data generation tools can improve robustness.

Finally, in intelligent digital health systems, it is more important than usual to provide reasons and explanations on predictions made by a machine learning system, so that interventions can be safely applied. For example, in \figref{importance_figures}d, we present a machine learning system that can detect abnormal blood glucose events such as hyperglycemia and hypoglycemia. However, detection alone may not be sufficiently insightful to prevent abnormal events because actionable feedback can provide with model-based reasoning or insightful feedback about how to prevent an undesired outcome. An explanation with a counterfactual explanation or alternative scenarios with proposed changes to improve a health outcome can overcome that limitation. Hence, it is important for an intelligent digital health system to be able to provide meaningful explanations.

Significant progress has been made in applying machine learning and artificial intelligence to healthcare challenges, addressing issues such as sensor failure, treatment adherence, neonatal risk prediction, and physical activity forecasting. These efforts strive to bridge the gap between clinical requirements and computational methods by integrating advanced deep learning techniques with domain-specific insights to improve the reliability, personalization, and explainability of health monitoring systems.

One study \cite{mamun2022designing} addresses the issue of missing data in sensor-based systems caused by sensor malfunctions or power limitations. By proposing an algorithm to reconstruct missing input data, the research demonstrated significant improvements in activity classification accuracy on benchmark datasets like MotionSense \cite{Malekzadeh:2018:PSD:3195258.3195260, Malekzadeh:2019:MSD:3302505.3310068}, MobiAct \cite{vavoulas2016mobiact}, and MHEALTH \cite{banos2014mhealth}. This contribution underscores the importance of building robust systems capable of maintaining reliable performance in real-world scenarios.

In the field of neonatal health, related work \cite{mamun2023neonatal,mamun2024use} explores the use of deep learning combined with counterfactual explanations to predict adverse labor outcomes. This approach addresses challenges such as class imbalance and explainability, aiming to reduce risks during childbirth and improve neonatal health outcomes. Similarly, research on treatment adherence prediction demonstrates the effectiveness of personalized models in forecasting adherence patterns, providing a framework for designing timely interventions to support patient care.

Efforts in multimodal data integration have also gained attention, particularly in the context of adaptive lifestyle interventions. By combining data from wearable devices, app engagement, and dietary habits, these studies \cite{mamun2022multimodal, mamun2024multimodal} enable the prediction of physical activity levels and postprandial glucose trends. The integration of explainable AI techniques and counterfactual reasoning in these systems supports informed decision-making for both users and healthcare providers.

While several prior review articles have surveyed different aspects of trustworthy AI in isolation, there remains a gap in systematically analyzing both robustness and explainability dimensions together in the context of digital health. This review uniquely bridges that gap by providing a comprehensive discussion of robustness and explainability, grounded in the practical demands of healthcare systems. It also critically examines existing evaluation metrics and proposes a taxonomy to guide future research at the intersection of reliable and explainable AI in digital health applications. This paper aims to balance breadth and depth by offering a concise overview of trustworthy AI, with a detailed focus on two key aspects: robustness and explainability.

\section{Background}
\subsection{Prior Reviews on Trustworthy AI}
The landscape of trustworthy AI research has been significantly enriched by numerous review efforts. Kaur et al. \cite{kaur2022trustworthy} provide a comprehensive analysis of the diverse requirements for trustworthy AI, including fairness, explainability, accountability, and reliability, offering insights into approaches to mitigate risks and improve societal acceptance. Kumar et al. \cite{kumar2020trustworthy} emphasize the ethical foundations of trustworthy AI, detailing how ethics can be embedded in system design and development, with a focus on practical applications in smart cities. Similarly, Kaur et al. \cite{kaur2021requirements} consolidate approaches for trustworthy AI based on the European Union's principles, presenting a structured overview for achieving reliable systems. Liu et al. \cite{liu2023trustworthy} delve into the alignment of large language models (LLMs), identifying key dimensions of trustworthiness such as safety, fairness, and adherence to social norms, while also conducting empirical evaluations to highlight alignment challenges. Lastly, Fehr et al. \cite{fehr2024trustworthy} assess the transparency of CE-certified \cite{ce_marking} medical AI products in Europe, revealing significant documentation gaps and calling for stricter legal requirements to ensure safety and ethical compliance in medical AI. These works collectively advance the understanding and implementation of trustworthy AI across various domains.

Continuing the exploration of trustworthy AI, Li et al. \cite{li2023trustworthy} propose a unified framework that integrates fragmented approaches to AI trustworthiness across the system lifecycle, addressing challenges such as robustness, fairness, and privacy preservation. Their comprehensive guide identifies actionable strategies for practitioners and policymakers to enhance the trustworthiness of AI systems. In the healthcare sector, Albahri et al. \cite{albahri2023systematic} present a systematic review of the trustworthiness and explainability of AI applications, highlighting the significant transparency and bias risks present in current methodologies. They propose guidelines and a detailed taxonomy for integrating explainable AI (XAI) methods into healthcare systems. Verma et al. \cite{verma2020counterfactual} focus specifically on counterfactual explanations, offering a rubric to evaluate counterfactual algorithms and identifying promising research directions for this critical aspect of model explainability. Saraswat et al. \cite{saraswat2022explainable} expand on the role of XAI in Healthcare 5.0, providing a solution taxonomy and case studies that demonstrate the potential of explainable AI techniques in improving operational efficiency and patient trust through privacy-preserving federated learning frameworks. Lastly, Band et al. \cite{band2023application} critically review XAI methodologies such as SHAP \cite{lundberg2017shap}, LIME \cite{ribeiro2016should}, and Grad-CAM \cite{selvaraju2017grad}, emphasizing their applicability to healthcare challenges like tumor segmentation and disease diagnosis. They offer a synthesis of usability and reliability studies, underscoring the transformative potential of XAI in medical decision-making. The review paper by Ojha et al. (2025) \cite{ojha2025navigating} explores the critical role of uncertainty in AI for healthcare, emphasizing the need for improved uncertainty quantification methods and addressing the gaps in understanding user perceptions of uncertainty and trustworthiness. Together, these works deepen our understanding of the pathways to achieving trustworthy and explainable AI systems, particularly in high-stakes domains like healthcare. A brief summary of the above papers is presented in \tblref{prior_reviews}.

\begin{table}[htbp]
\centering

\scalebox{0.8}{\begin{tabular}{llp{8cm}}
\hline
\textbf{Publication} & \textbf{Application} & \textbf{Brief Description} \\ \hline
Kumar et al. (2020) \cite{kumar2020trustworthy} & Smart Cities & Highlights the embedding of ethical foundations in AI system design and development, focusing on practical applications in smart cities. \\ \hline
Verma et al. (2020) \cite{verma2020counterfactual} & Counterfactual Explanations & Offers a rubric to evaluate counterfactual algorithms, identifying research directions for this critical aspect of model explainability. \\ \hline
Kaur et al. (2021) \cite{kaur2021requirements} & General AI systems & Consolidates approaches for trustworthy AI based on the European Union's principles, presenting a structured overview for achieving reliable systems. \\ \hline
Kaur et al. (2022) \cite{kaur2022trustworthy} & General AI systems & Provides a comprehensive analysis of the requirements for trustworthy AI, including fairness, explainability, accountability, and reliability, with approaches to mitigate risks and improve societal acceptance. \\ \hline
Saraswat et al. (2022) \cite{saraswat2022explainable} & Healthcare 5.0 \cite{mbunge2021sensors} & Provides a taxonomy and case studies demonstrating the potential of explainable AI in improving operational efficiency and patient trust using privacy-preserving federated learning frameworks. \\ \hline
Liu et al. (2023) \cite{liu2023trustworthy} & Large Language Models (LLMs) & Explores the alignment of LLMs with safety, fairness, and social norms, providing empirical evaluations to highlight alignment challenges. \\ \hline
Band et al. (2023) \cite{band2023application} & Medical Decision-Making & Critically reviews XAI methodologies like SHAP \cite{lundberg2017shap}, LIME \cite{ribeiro2016should}, and Grad-CAM \cite{selvaraju2017grad}, emphasizing their usability and reliability for applications like tumor segmentation and disease diagnosis. \\ \hline
Albahri et al. (2023) \cite{albahri2023systematic} & Healthcare AI & Systematically reviews trustworthiness and explainability in healthcare AI, highlighting transparency and bias risks and proposing a taxonomy for integrating XAI methods into healthcare systems. \\ \hline
Li et al. (2023) \cite{li2023trustworthy} & General AI systems & Proposes a unified framework integrating fragmented approaches to AI trustworthiness, addressing challenges such as robustness, fairness, and privacy preservation. \\ \hline
Fehr et al. (2024) \cite{fehr2024trustworthy} & Medical AI in Europe & Assesses the transparency of CE-certified medical AI products, revealing significant documentation gaps and calling for stricter legal requirements to ensure safety and ethical compliance. \\ \hline

Ojha et al. (2024) \cite{ojha2025navigating} & Healthcare AI & Focuses on one specific area of trustworthiness, that is, uncertainty. \\ \hline

\textit{\textbf{This paper}} & Digital Health & Provides a comprehensive review and discussion of the aspects of trustworthy AI in digital health systems, with a focus on robustness and explainability. \\ \hline

\end{tabular}}
\caption{Prior review works in the area of Trustworthy AI and the contribution of this paper.}
\label{tbl:prior_reviews}
\end{table}

\subsection{Branches of Trustworthy AI}
Trustworthy AI can be categorized into five interrelated branches, each addressing distinct aspects of ethical and reliable AI systems. Robustness ensures that systems perform reliably even under varied or adverse conditions, such as noisy inputs or missing data. Explainability makes AI models and their decisions understandable, allowing users to comprehend and trust the decisions made by the models. Fairness focuses on preventing biases and ensuring equitable outcomes for all users. Privacy safeguards sensitive user data throughout the AI lifecycle, maintaining confidentiality and security. Finally, accountability establishes mechanisms for responsibility and oversight, ensuring that AI systems operate within ethical and legal boundaries. Together, these branches provide a comprehensive framework for building AI systems that inspire trust and confidence among users.

\subsection{Challenges in Different Phases of a Machine Learning System}
\begin{figure}
    \centering
    \includegraphics[width=0.98\linewidth]{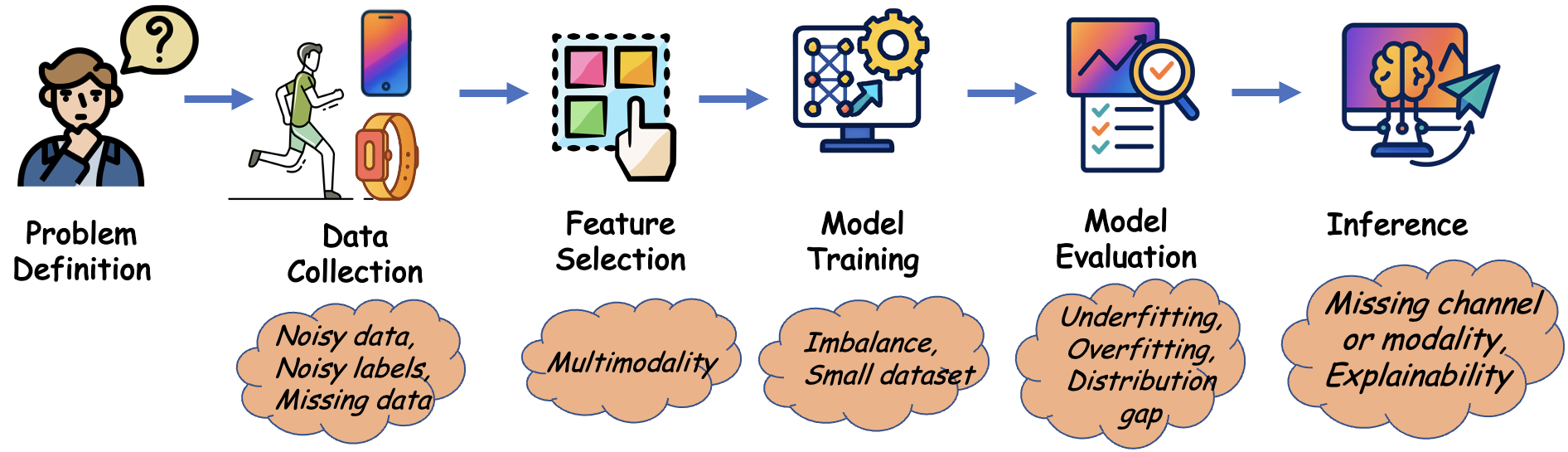}
    \caption{Challenges in machine learning systems in different phases of the development lifecycle.}
    \label{fig:phase_and_challenge}
\end{figure}

Challenges in Trustworthy AI arise at various stages of the machine learning pipeline. During the problem definition phase, it is essential to clearly outline the scope and objectives while accounting for ethical and societal implications. Data collection often involves issues such as noisy data, noisy labels, and missing data, which can compromise the quality of training datasets. In feature selection, managing multimodality becomes crucial to effectively integrate heterogeneous data sources. The model training phase encounters difficulties like class imbalance, small datasets, and the risk of overfitting, while model evaluation must address underfitting, overfitting, and performance gaps due to distribution shifts. Finally, the inference stage brings challenges such as missing channels or modalities and ensuring explainability during deployment. Addressing these challenges holistically is key to creating robust and trustworthy AI systems. We present an overview of these challenges in \figref{phase_and_challenge}.

\subsection{Designing Robust Machine Learning Models}

Robustness is a fundamental aspect of Trustworthy AI, ensuring that models maintain reliable performance across a wide range of conditions \cite{europaEthicsGuidelines, li2023trustworthy}. For noisy data, noisy labels, and missing data, techniques such as denoising autoencoders \cite{vincent2008extracting}, convolutional networks \cite{mamun2022designing}, and fuzzy c-means clustering \cite{aydilek2013hybrid} have proven effective in reconstructing or imputing missing information. Masked autoencoders \cite{he2022masked}, for instance, use random masking and self-supervised learning to predict and reconstruct missing patches in input data, enhancing model reliability. Robustness also extends to multimodal data integration and addressing class imbalance. Multimodal deep learning methods combine data from diverse sources, improving system performance and resilience. Data balancing techniques, such as re-weighting or re-sampling, help tackle class imbalance issues, especially in small datasets \cite{zhai2022binary, mamun2024use, shah2025enhancing}.

Several contributions highlight advancements in robustness. For example, systems designed to mitigate sensor failure or system designed to be able to operate with low energy \cite{mamun2022designing, hussein2023cim, hussein2023energy}. Robustness for noisy labels has improved activity recognition in free-living environments \cite{soumma2025domain, alinia2020actilabel, mirzadeh2019labelmerger}, while robustness for multimodal data has applications in activity forecasting \cite{mamun2022multimodal}. Additionally, addressing class imbalance has led to innovations in neonatal health applications, reducing overfitting and improving outcomes in small datasets. A summary of different contributions in the field developing robust machine learning solutions is provided in \tblref{robustness1} and \tblref{robustness2}. 

\section{Application-Specific Trust Concerns}
While general principles of trustworthy AI, such as fairness, explainability, and robustness, remain applicable across domains, their manifestation in healthcare is often domain-specific. This section explores how trust concerns arise and are addressed within distinct areas of digital health. By contextualizing trust within each domain, we highlight the nuances and practical challenges of building AI systems that clinicians, patients, and regulators can rely on.

\subsection{Radiology and Medical Imaging}

Medical imaging has been a fertile ground for deep learning-based diagnostics \cite{bernal2017comparative, zhou2019unet++, hasani2022trustworthy}. However, black-box models pose concerns regarding explainability and safety. Trustworthiness in this domain has been approached through saliency maps, Grad-CAM \cite{selvaraju2017grad}, counterfactual visualization \cite{thiagarajan2022training}, and attention-based explainability \cite{chung2024evaluating}. Techniques like domain adaptation and test-time augmentation are used to enhance robustness and consistency \cite{guan2021domain, wang2019aleatoric}.

\subsection{Cardiovascular Health}
AI in cardiology is used in arrhythmia detection and atrial fibrillation prediction from ECG signals \cite{ansari2023deep, guan2024interpretable}. Trust hinges on robust signal modeling, explainability, and alignment with clinical workflows.  Wearable ECG applications demand calibrated outputs and minimal false alarms to avoid unnecessary interventions.

Wearables generate continuous data streams for health tracking, yet present challenges in sensor fidelity, environmental noise, and user variability \cite{soumma2025domain}. Trustworthy modeling in this domain emphasizes online learning, anomaly detection, as well as designing a system so that it can handle unexpected scenarios such as sensor failure or unavailability during inference \cite{mamun2022designing}.

\subsection{Metabolic Health}

Applications in metabolic health include blood glucose forecasting \cite{daniels2021multitask, farahmand2025attengluco, mamun2025llmpoweredpredictionhyperglycemiadiscovery}, lifestyle personalization \cite{arefeen2024glyman, arefeen2025glytwin}, and insulin resistance monitoring. These models typically integrate data from CGMs, diet logs, and wearables \cite{mamun2025llmpoweredpredictionhyperglycemiadiscovery}. Key challenges include user variability, missing data, class imbalance, and context sensitivity. Explainability techniques, e.g., counterfactual explanations and privacy-preserving learning frameworks (e.g., federated learning) are often employed to enhance trust \cite{arefeen2023designing, arefeen2024glyman, mamun2025llmpoweredpredictionhyperglycemiadiscovery, chellamani2025diabetes, shah2025enhancing}.

\subsection{Neonatal Health and Pediatrics}

AI in pediatric and neonatal care faces the dual challenge of data scarcity and high stakes \cite{shazly2022impact, pruksanusak2022comparison, mamun2023neonatal, mamun2024use}. Explainability is crucial, as caregivers and providers demand transparency for decisions affecting vulnerable populations. Methods include what-if analysis, causal feature attribution, and generalizability testing across institutions \cite{mamun2024use}.

\subsection{Mental Health and Addiction Recovery}
AI systems are increasingly used to monitor mental health and support addiction recovery in free-living environments. To address the challenge of limited labels in substance use detection, recent frameworks leverage self-supervised learning on wearable sensor data, enabling accurate detection of cannabis use with minimal supervision \cite{azghan2025cudle}. Stress detection in individuals recovering from Alcohol Use Disorder has similarly benefited from optimized sensor selection and context modeling, revealing skin conductance as a key signal \cite{sah2023stress}. A broader review of ML in addiction research highlights the use of brain imaging, behavioral phenotyping, and ensemble models like random forests for early screening and monitoring \cite{chhetri2023machine}. These approaches emphasize trust through robust learning under label scarcity, personalization, and clinically relevant explanations.

\subsection{Brain Health}

AI methods for brain health, especially neurodegenerative diseases like Alzheimer's and Parkinson's face distinct trust challenges due to the sensitive, subjective, and progressive nature of these conditions \cite{li2023early, wu2024wearable, mostafavi2025detection}. In brain health, models using mobile sensing and behavioral data raise privacy concerns and benefit from causal modeling and clinician-in-the-loop systems.  To ensure safety and transparency, approaches such as federated learning, uncertainty estimation, and saliency-based explanations are being adopted \cite{soumma2024freezing}.

\subsection{Intensive Care and Monitoring}

AI is increasingly used in ICUs for early prediction of sepsis, patient deterioration, and resource management. A deep learning model trained on over 130{,}000 ICU admissions across multiple countries demonstrated strong generalization and detected sepsis 3.7 hours before onset \cite{moor2023predicting}. External validation and low false alarm rates contributed to clinical trust. Reviews emphasize the importance of explainability, ethical safeguards, and robust modeling under data variability \cite{stylianides2025ai}.

\subsection{Public Health and Epidemiology}

In public health, AI supports disease surveillance, epidemic forecasting, and community-level interventions. Studies highlight the promise of citizen science, predictive analytics, and participatory tools to improve health equity \cite{king2025promise, olawade2023using}. Responsible AI frameworks are essential to address fairness, transparency, and ethical risks in real-time surveillance systems \cite{borda2022ethical}. Trust relies on inclusive design, societal accountability, and transparent public engagement.

\section{Advances in Trustworthy AI for Digital Health}

\subsection{Label Scarcity and Data-Efficient Learning}

Trustworthy AI in digital health emphasizes the development of robust and explainable systems to address unique challenges in healthcare. For example, label scarcity in free-living environments complicates supervised learning for detecting health-related behaviors. The CUDLE framework addresses this by leveraging self-supervised learning to identify cannabis consumption moments using wearable sensors. By training on augmented data for robust feature extraction and fine-tuning with minimal labeled data, CUDLE achieves higher accuracy compared to traditional supervised methods, even with 75\% fewer labels \cite{azghan2025cudle}. Similarly, clinical speech AI must account for limited data availability and overfitting risks. Integrating insights from speech science, explainable models, and robust validation frameworks can mitigate these issues and accelerate the translation of speech biomarkers for clinical use \cite{berisha2024responsible}. These approaches demonstrate the importance of data-efficient and explainable methodologies in healthcare AI.

\subsection{Forecasting and Personalized Interventions}

Probabilistic forecasting in public health further underscores the need for trustworthy AI. During the COVID-19 pandemic, ensemble models synthesizing predictions from multiple groups consistently outperformed individual models in forecasting mortality rates. This collaborative effort highlighted the importance of active coordination between public health agencies, academia, and industry for developing reliable modeling capabilities under real-world constraints \cite{cramer2022evaluation}. In digital health interventions, personalized approaches such as smartphone and text-message-based systems for managing type 2 diabetes showed significant improvements in glycemic control compared to website-based interventions \cite{watterson2018improved}. The higher reach and uptake of these modalities suggest their practicality for real-world deployment, though optimizing delivery mechanisms remains crucial \cite{moschonis2023effectiveness}.

\subsection{Self-Supervised Learning and Cross-Domain Generalization}

Self-supervised learning also plays a transformative role in medical AI \cite{azghan2025cudle}, enabling models to learn from large-scale unannotated data across diverse modalities such as medical images and bioelectrical signals \cite{huang2023self, ouyang2022self}. These methods address challenges like limited annotated datasets and biased data collection, facilitating the development of scalable and generalizable AI systems \cite{krishnan2022self}. Additionally, inherent sensor redundancies have been exploited to enhance anomaly detection in sensor-based systems, demonstrating the potential of cross-domain techniques for improving trustworthiness and robustness in digital health applications \cite{he2020exploring}. Together, these advancements highlight a growing emphasis on creating AI systems that are not only accurate but also ethical, transparent, and adaptable to diverse healthcare contexts.

\subsection{Robustness and Clinical Utility}

Several additional works have addressed challenges in trustworthy AI for digital health, such as sensor failure and imbalanced datasets in healthcare applications. For example, efforts in AIMEN \cite{mamun2024use} and medication adherence forecasting \cite{mamun2025aimi} have focused on building robust, explainable systems that aim to improve reliability and clinical utility, particularly in the contexts of neonatal health and treatment adherence. These works contribute to enhancing the trustworthiness and effectiveness of AI applications in healthcare.

\subsection{Scientific Discovery and Human Oversight}

A recent example of progress toward trustworthy AI is Google's Co-Scientist system \cite{gottweis2025towards}, which introduces a multi-agent framework for automated scientific discovery while maintaining human oversight. Designed to generate, critique, and refine hypotheses grounded in literature, the system exemplifies transparency and accountability through its debate-style evaluation and citation-based reasoning. Importantly, the inclusion of scientists in the loop ensures explainability and alignment with domain knowledge, supporting responsible deployment in high-stakes fields such as biomedical research. While broader ethical considerations remain underexplored, the Co-Scientist represents a promising step toward collaborative, robust, and explainable AI in scientific and healthcare contexts.

\begin{table}[h!]
\centering
\scalebox{0.7}{\begin{tabular}{p{4.5cm}llp{4.5cm}p{4.5cm}}
\hline
\textbf{Name} & \textbf{Authors} & \textbf{Year} & \textbf{Method} & \textbf{Original Application} \\ \hline
Denoising Autoencoder \cite{vincent2008extracting} & Vincent et al. &  2008 & Uses noisy data as input and the corresponding clean data as output to train & Denoising images \\ \hline
Masked Autoencoder \cite{he2022masked} & He et al. & 2021 & The positions of the missing patches are used to improve reconstruction. & Recreates missing patches in images \\ \hline
Generalized Butterworth Filter \cite{selesnick2002generalized} & Selesnick and Burrus & 1998 & A Butterworth filter is a signal filter with a maximally flat passband response, minimizing ripples and ensuring smooth attenuation to the stopband. & Reduces noise in time-series data \\ \hline
Missing data imputation with Fuzzy c-means clustering \cite{aydilek2013hybrid} & Aydilek and Arslan & 2013 & A hybrid approach combines fuzzy c-means clustering, support vector regression, and a genetic algorithm to estimate missing values. & Improves imputation performance, outperforming zero imputation and other traditional methods. \\ \hline
ActiLabel \cite{alinia2020actilabel} & Alinia et al. & 2020 & Dependency graphs to capture structural similarities and map activity labels between domains. & Improve activity recognition’s usability and performance \\ \hline
Missing Sensor Data Reconstruction Algorithm \cite{mamun2022designing} & Mamun et al. & 2022 & Proposes an algorithm to reconstruct missing input data in sensor-based health monitoring systems, improving prediction accuracy on multiple activity classification benchmarks. & Reconstructing missing sensor data \\ \hline
CIM: Clustering-based Energy-Efficient Data Imputation Method \cite{hussein2023cim} & Hussein and Bhat & 2023 & CIM detects missing sensors, predicts their clusters for imputation using a mapping table, and determines activities through imputation-aware classification or a reliable activity classifier. & Detecting and imputing missing sensor data \\ \hline

Cross-Domain Conditional Diffusion Models for
Time Series Imputation \cite{zhang2025cross} & Zhang et al. & 2025 & Introduces a diffusion-based method for cross-domain time series imputation that handles domain shifts and missing data via spectral interpolation and consistency alignment. & Time series data \\ \hline
\end{tabular}}
\caption{Summary of robustness methods for missing data, noisy data, and related challenges and their applications.}
\label{tbl:robustness1}
\end{table}

\begin{table}[h!]
\centering
\scalebox{0.7}{\begin{tabular}{p{4.5cm}llp{4.5cm}l}
\hline
\textbf{Name} & \textbf{Authors} & \textbf{Year} & \textbf{Method} & \textbf{Data Type} \\ \hline
Multimodal deep learning \cite{ngiam2011multimodal} & Ngiam et al. &  2011 & Bimodal deep autoencoder and Restricted Boltzmann Machine to outperform unimodal classifiers & Video, Audio \\ \hline
SMOTE \cite{chawla2002smote} & Chawla et al. &  2002 & K-Nearest-neighbor based synthetic data generation & Tabular, Transformed features \\ \hline
AdaSYN \cite{he2008adasyn} & He et al. &  2008 & Synthetic data generator with higher priority near the decision boundary & Tabular, Transformed features \\ \hline
CTGAN \cite{xu2019modeling} & Xu et al. &  2019 & A modified GAN for tabular data with different processing for categorical and numerical features & Tabular \\ \hline
Binary Imbalanced Data Classification \cite{zhai2022binary} & Zhai et al. & 2021 & GAN and discarding of batch data based on silhouette score & Tabular \\ \hline
AIMEN and R-AIMEN \cite{mamun2024use} & Mamun et al. & 2024 & CTGAN based data balancing with or without restrictions on similarity and type of the generated data & Tabular \\ \hline
MetaBoost \cite{shah2025enhancing} & Shah et al. & 2025 & Hybrid data balancing method that creates batches of synthetic data from weighted combinations of multiple balancing methods & Tabular \\ \hline
Dropout \cite{srivastava2014dropout} & Srivastava et al. & 2014 & Randomly disables a number of neurons of the previous layer during training time. In the inference time, the weights are adjusted to maintain consistency. & Any data type \\ \hline
Knowledge-guided transformer for forecasting \cite{qi2021known} & Qi et al. & 2021 & Uses future knowledge (future promotions) to improve performance of forecasting & Time-series \\ \hline
\end{tabular}}
\caption{Summary of robustness methods for multimodality, class imbalance, overfitting, and applications.}
\label{tbl:robustness2}
\end{table}

\section{Explainability in Machine Learning}
The growing importance of explainability in machine learning (ML) and artificial intelligence (AI) has led to the development of several methods to make complex models more transparent and their predictions understandable to users. These techniques, ranging from feature importance scores to counterfactual explanations, help bridge the gap between high-performing but opaque models and their practical, responsible deployment in real-world applications, especially in critical fields like healthcare. A summary of different explainable AI methods is provided in \tblref{explanation_methods}. Below, we review various explainable AI approaches, with a particular emphasis on counterfactual explanations and their applications.

\subsection{Explainable AI Methods in ML Models}
Explainable AI focuses on making machine learning models transparent and ensuring that users can understand the reasoning behind predictions. Methods such as Shapley values \cite{shapley1953value} provide feature importance scores, which help users discern the factors influencing model predictions. Another prominent technique is counterfactual explanations, which explore how altering input features would lead to different outcomes, offering actionable insights for decision-making. Tools like DiCE \cite{mothilal2020explaining} generate diverse counterfactuals, enhancing both explainability and usability in practical settings. Further advancements include NICE (Nearest Instance Counterfactual Explanations) \cite{brughmans2024nice}, which produces intuitive and transparent explanations by finding the nearest unlike neighbor and then modifying it. These innovations have been applied in critical domains, such as neonatal health, where NICE and SHAP were used to explain health predictions \cite{mamun2024use}, offering clinicians a better understanding of risk factors and outcomes. Similarly, DiCE and other methods for counterfactual explanations have been applied to multimodal hyperglycemia prediction, improving interpretability and reducing the risks associated with glucose management \cite{mamun2025llmpoweredpredictionhyperglycemiadiscovery, arefeen2024glyman}.

\subsection{Concept and Visual Explanations}

In the realm of explainable AI (XAI), recent work has focused on enhancing the depth and quality of solutions through both local and global explanations. For instance, Achtibat et al. \cite{achtibat2023attribution} propose Concept Relevance Propagation (CRP), a novel method that bridges the gap between local and global XAI approaches. By addressing both the "where" and "what" aspects of model predictions, CRP provides human-understandable explanations via techniques like concept atlases and subspace analysis. Similarly, Dreyer et al. \cite{dreyer2023revealing} extend CRP to segmentation and object detection tasks with L-CRP, helping locate and visualize relevant learned concepts and uncovering biases in models like DeepLabV3+ \cite{chen2018encoder} and YOLOv6 \cite{li2022yolov6}. Saliency maps \cite{simonyan2013deep} and deconvnets \cite{zeiler2014visualizing} were popular technique for visualizing convolutional networks. Amorim et al. \cite{amorim2023evaluating} developed a framework that assesses the faithfulness of saliency maps, focusing on tumor localization in medical images, while Chattopadhay et al. \cite{chattopadhay2018grad} introduced Grad-CAM++, an improved saliency map method for better localization of multiple object occurrences in images. These approaches underscore the importance of improving visual explanations to provide more transparent and explainable models.

\subsection{Regularization and Novel Frameworks for Model Robustness}

Regularization techniques are pivotal in enhancing model robustness and explainability. Choi et al. \cite{choi2020role} explored the impact of orthogonality constraints on feature representations in deep learning models. Their work introduced an Orthogonal Sphere regularizer, which improves feature diversity, enhances semantic localization, and reduces calibration errors across various datasets. Concurrently, Du et al. \cite{du2025haloscope} tackled the issue of hallucinations in large language models (LLMs) with their HaloScope framework. This method uses unlabeled LLM generations to train a classifier to detect hallucinations, achieving superior performance without the need for additional data collection. Both contributions highlight the critical role of structural constraints and novel frameworks in enhancing the explainablity and reliability of AI systems.

\begin{table}[h!]
\centering

\scalebox{0.7}{\begin{tabular}{p{4.5cm}llp{4.5cm}l}
\hline
\textbf{Name} & \textbf{Authors} & \textbf{Year} & \textbf{Method} & \textbf{Data Type} \\ \hline
LIME \cite{ribeiro2016should} & Ribeiro et al. & 2016 & Surrogate interpretable model for estimating effect. & Tabular, Text, Image \\ 
Shapley \cite{shapley1953value} & L. Shapley & 1953 & Measure effect of a feature on different coalitions of all other features. & Tabular, Time-series, Image \\ 
SHAP \cite{lundberg2017shap} & Lundberg and Lee &  2017 & Efficient approximation based on Shapley (practical when the number of features is large). & Tabular, Time-series, Image \\ 
GradCAM \cite{selvaraju2017grad} & Selvaraju et al. & 2017 & Generates class-specific heatmaps by calculating the gradients of the target class w.r.t. feature maps. & Image \\ 
Layer-wise Relevance Propagation \cite{binder2016layer} & Binder et al. & 2016 & Assigns relevance scores to input features by propagating the output decision backwards. & Image, Tabular, Text \\

Integrated Gradients \cite{sundararajan2017axiomatic} & Sundararajan et al. & 2017 & A path-based attribution method that assigns feature importance by accumulating gradients from a baseline to the input. & Image \\ 

NICE \cite{brughmans2024nice} & Brughmans et al. &  2023 & Counterfactual explanation based on a modified nearest unlike neighbor. & Tabular \\ 
DiCE \cite{mothilal2020explaining} & Mothilal et al. &  2020 & Counterfactual explanations that optimize on validity, proximity, diversity, and sparsity. & Tabular \\ 
Semi-factual explanation \cite{aryal2024semi} & Aryal and Keane & 2024 & Finds alternate feature values on the same class. Can be counterfactual (CF)-free or CF-guided. & Tabular \\

Multi-Objective Counterfactuals \cite{dandl2020multi} & Dandl et al. & 2020 & Counterfactual method satisfying 4 objective functions: validity, distance, sparsity, distribution sanity. & Tabular \\ \hline
\end{tabular}}
\caption{An overview of a few different explainable AI methods.}
\label{tbl:explanation_methods}
\end{table}

\subsection{Counterfactual Explanations in XAI}

Counterfactual explanations have become a focal point of research in XAI, as they help clarify what changes to the input data could lead to different model predictions. Aryal \cite{aryal2024semi} introduces the concept of semi-factual explanations, which highlight input changes that do not alter the model’s output. These explanations offer insights into the stability of model predictions, thus enhancing trust and interpretability. In parallel, Bajaj et al. \cite{bajaj2021robust} have worked on generating robust counterfactual explanations for Graph Neural Networks (GNNs), ensuring that the explanations remain stable even in noisy environments. Their method identifies subgraphs that strongly influence predictions while maintaining human interpretability. Brughmans et al. \cite{brughmans2024nice} have developed NICE, a tool designed to produce counterfactuals for tabular data, optimizing for sparsity, proximity, and plausibility, thus ensuring that the explanations are both efficient and applicable to a wide range of models.

\subsubsection{Benchmarking and Frameworks for Counterfactuals}

Several studies have also focused on benchmarking and unifying counterfactual explanation methods. Guidotti \cite{guidotti2024counterfactual} provides a comprehensive review and benchmarking of counterfactual explanation techniques, categorizing them based on properties such as stability, diversity, and actionability. This analysis highlights trade-offs in counterfactual generation and calls for methods that can balance multiple desirable properties. These surveys and frameworks are essential for refining counterfactual techniques and ensuring they are actionable, efficient, and interpretable for both technical and non-technical users.

\subsubsection{Applications of Counterfactual Explanations in Healthcare}

Counterfactual explanations play a vital role in healthcare applications, particularly in systems like AIMEN \cite{mamun2024use} and GlucoLens \cite{mamun2025llmpoweredpredictionhyperglycemiadiscovery}. These systems use counterfactual reasoning to provide actionable insights for clinicians and patients, such as demonstrating the effects of alternative interventions or behavioral changes. By showing how small changes in input features could lead to different outcomes, counterfactual explanations not only improve the interpretability of AI-driven decisions but also enhance user trust in these systems. The use of counterfactuals in healthcare systems aligns with the ongoing need to make AI both more transparent and user-friendly, particularly in high-stakes domains such as healthcare and clinical decision-making.

\section{Explainable AI Techniques}

\subsection{LIME (Local Interpretable Model-agnostic Explanations)}
\begin{figure}[h]
    \centering
    \includegraphics[width=0.85\linewidth]{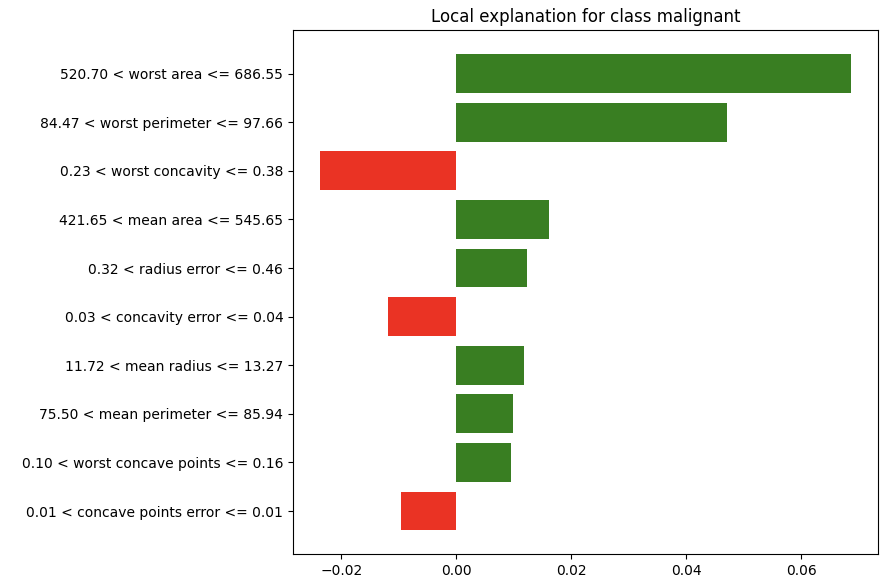}
    \caption{Example of a LIME plot for breast cancer detection. In this example, we applied LIME on the publicly available breast cancer dataset \cite{breast_cancer_wisconsin_diagnostic_17, scikitlearnLoad_breast_cancer}.}
    \label{fig:lime_example}
\end{figure}

LIME \cite{ribeiro2016should} provides explanations for predictions by fitting interpretable models (e.g., linear regression) locally around the data point of interest. It perturbs the input data to create a neighborhood and observes the behavior of the black-box model. The resulting simplified model highlights the contribution of individual features, offering an easy-to-understand explanation of the prediction. LIME’s flexibility allows it to work with any machine learning model, making it widely applicable. An example of a LIME plot is presented in \figref{lime_example}.

\subsection{Shapley Values}
Shapley values \cite{shapley1953value} are derived from cooperative game theory and represent a method to fairly distribute a total value among contributors (features in machine learning). They calculate the marginal contribution of a feature by averaging over all permutations of feature inclusion. While computationally intensive, Shapley values form the theoretical basis for tools like SHAP, ensuring fairness and consistency in feature attribution.

\subsection{SHAP (SHapley Additive exPlanations)}
\begin{figure}[h]
    \centering
    \includegraphics[width=0.90\linewidth]{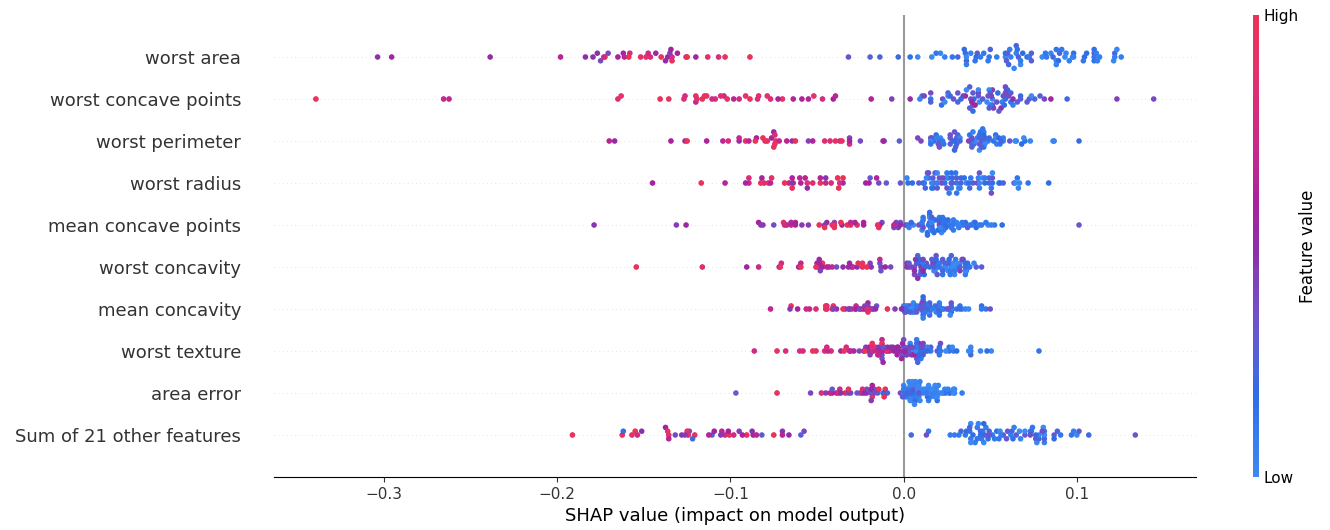}
    \caption{Example of a SHAP BeeSwarm plot for breast cancer detection. We applied SHAP on the breast cancer dataset \cite{breast_cancer_wisconsin_diagnostic_17, scikitlearnLoad_breast_cancer}.}
    \label{fig:shap_example}
\end{figure}

SHAP \cite{lundberg2017shap} employs concepts from cooperative game theory, specifically Shapley values, to quantify the contribution of each feature to a model’s prediction. By averaging over all possible feature subsets, SHAP guarantees a fair distribution of the prediction value among the features. Its additive nature makes it compatible with both linear and complex models, providing consistent and theoretically sound explanations. An example of a SHAP plot is presented in \figref{shap_example}.

\subsection{LRP (Layer-wise Relevance Propagation)}
LRP \cite{binder2016layer} is a technique for explaining neural networks by redistributing the model's output backward through the layers to the input. Using relevance propagation rules, it assigns relevance scores to each neuron and feature, indicating their contribution to the prediction. LRP is particularly useful for understanding the decision-making process in deep networks like CNNs and RNNs, especially in image and text tasks.

\subsection{GradCAM (Gradient-weighted Class Activation Mapping)}
\begin{figure}[h]
    \centering
    \includegraphics[width=0.95\linewidth]{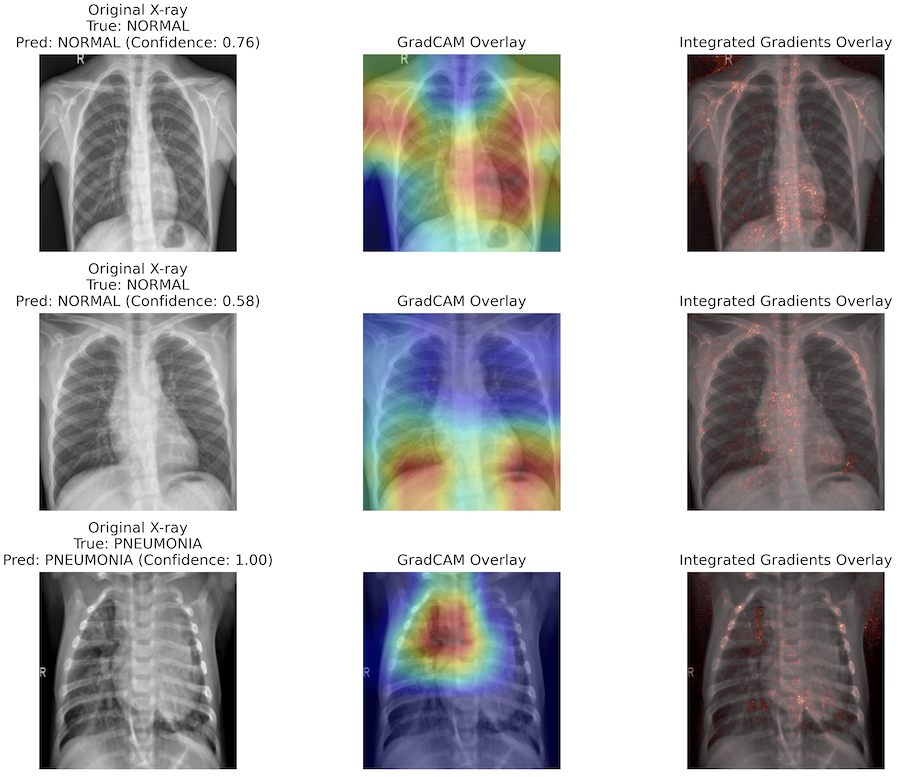}
    \caption{Example of GradCAM and Integrated Gradients plots for detecting pneumonia from chest X-rays from a publicly available dataset \cite{kermany2018identifying}. In this example, we trained a ResNet18 model that achieves a test accuracy of 82\% and a macro average test F1 score of 0.77. Then, GradCAM and Integrated Gradients were applied for the explanation of the model's predictions. In the first normal case, the explanation methods highlight the overall X-ray as most of the X-ray suggests a normal finding. In the second example, the model predicts a normal finding correctly, but with lower confidence, and both explanation methods highlight the areas that led the model to predict normal. In the third example, the model correctly predicts pneumonia with high confidence, and GradCAM and Integrated Gradients refer to the areas responsible for this diagnosis, because of the possibility of reticular opacities, a common indicator of pneumonia.}
    \label{fig:gradcam_integrated_gradients}
\end{figure}

GradCAM \cite{selvaraju2017grad} generates visual explanations for CNN-based models by highlighting important regions in input images. It computes gradients of the model’s output with respect to the feature maps of the last convolutional layer, using these gradients to create heatmaps. The resulting heatmaps overlay the input image to show areas that strongly influence the model’s decision, making it highly intuitive for interpreting visual data. An example of the GRADCAM plot is presented in \figref{gradcam_integrated_gradients}.

\subsection{Integrated Gradients}
Integrated Gradients \cite{sundararajan2017axiomatic} is a widely used attribution technique designed to explain predictions made by deep neural networks in a theoretically grounded manner. It satisfies key axioms such as sensitivity and implementation invariance, which ensure that the attributions are consistent and faithful to the model’s behavior. The method estimates feature importance by tracing a straight-line path from a baseline input (often a zero or black image) to the actual input, and accumulating gradients along this path. This process results in attribution scores for each feature, capturing how changes in the input influence the model’s output. The resulting attribution map highlights the most influential input features, offering a more stable and interpretable explanation compared to basic gradient-based methods. This makes Integrated Gradients especially valuable in sensitive domains like medical imaging and other applications requiring transparent decision-making. An example visualization of this method is shown in \figref{gradcam_integrated_gradients}.

\subsection{NICE (Nearest Instance Counterfactual Explanations)}
NICE \cite{brughmans2024nice} focuses on generating counterfactual explanations by identifying the nearest instance in the feature space that results in a different prediction. It minimizes the distance between the original and counterfactual inputs while ensuring that the counterfactual belongs to a different class. This approach ensures that the generated explanations are not only interpretable but also closely aligned with the data distribution, making them realistic and actionable.

\subsection{DiCE (Diverse Counterfactual Explanations)}
DiCE \cite{mothilal2020explaining} generates counterfactual explanations to highlight how minimal changes to input features could alter a model’s prediction. It focuses on creating diverse explanations, providing multiple plausible alternatives for user exploration. By balancing proximity, diversity, and feasibility, DiCE is well-suited for use in sensitive domains like finance and healthcare, where interpretability is crucial. 

\subsection{CFNOW}
CFNOW \cite{de2024model} adopts a two-step optimization strategy for generating counterfactual explanations. The first step, counterfactual (CF) search, identifies an initial solution that alters the original classification, focusing on finding a feasible counterfactual without ensuring it is optimal. The second step, CF improvement, refines this initial solution using an objective function, such as minimizing the distance between the original and counterfactual instances, to enhance its quality. CFNOW supports four approaches: greedy simple (CGS), greedy optimized (CGO), random simple (CRS), and random optimized (CRO). While simple methods prioritize speed and computational efficiency, optimized methods refine the counterfactual for improved plausibility and adherence to specific objectives. CFNOW processes diverse data types, including tabular, image, and textual inputs, by converting them into tabular-like features. It supports binary and multiclass classification tasks with options to compare factual and counterfactual classes. This flexibility makes CFNOW a powerful tool for generating realistic and adaptable counterfactual explanations across a variety of domains.

\section{Trustworthy AI in the Era of LLMs}

Emerging technologies leveraging advancements in large language models (LLMs) are rapidly reshaping numerous domains. The advent of systems like ChatGPT has showcased the potential of LLMs to enhance research, education, and medical practices by democratizing access to information and improving decision-making capabilities \cite{clusmann2023future}. However, these advancements come with challenges, such as the risks of misinformation and ethical concerns related to accountability. Beyond healthcare, the development of Llama 3 models demonstrates strides in multilingual support, coding, and multimodal functionalities, enabling applications in diverse fields from content generation to knowledge synthesis \cite{dubey2024llama}. By integrating image, video, and speech processing, these systems push the boundaries of AI capabilities, further amplified by novel approaches like retrieval-augmented generation (RAG), which combines parametric and non-parametric memories to enhance specificity and factual accuracy in language generation \cite{lewis2020retrieval}.

The focus on reasoning capabilities has also gained momentum with models like DeepSeek-R1, which employs reinforcement learning to refine logical inference and problem-solving skills \cite{guo2025deepseek}. This approach complements the zero-shot reasoning capabilities explored in foundational works on chain-of-thought prompting, which reveal the untapped cognitive potential of LLMs \cite{kojima2022large}. Furthermore, specialized methods like RNAS-CL \cite{nath2024rnas} improve the robustness of neural architectures through cross-layer knowledge distillation, addressing challenges of adversarial vulnerability in AI systems \cite{nath2024rnas}. Together, these innovations underscore a transformative era where AI systems evolve to be not only versatile but also reliable and ethical, fostering advancements across healthcare, education, mental health, and beyond.

In our opinion, the explanations for any prediction or output by language models can be divided into three categories, i) the explanations are generated by the same language model \cite{chen2024models}, ii) the explanations are generated by a non-language model, iii) the explanations are generated by another language model \cite{singh2023explaining}. The previous publications usually divide the first category again into two subcategories \cite{chen2024models}: a) chain of thought (CoT) \cite{nye2021show, kojima2022large, wei2022chain}, or b) post-hoc \cite{camburu2018snli, park2018multimodal} .

As these technologies become integral to critical decision-making processes, ensuring their trustworthiness is paramount \cite{liu2023trustworthy, baker2023explainable}. In the context of LLMs, transparency refers to the ability to understand how these models arrive at their conclusions, a challenge given their complex, opaque nature. The recent reasoning models are being created in a way so that a detailed explanation of the steps are provided with the LLM response \cite{guo2025deepseek, jaech2024openai, bercovich2025llama, rastogi2025magistral}. It was found that providing reasonings behind the decision often improves the accuracy of the prediction of large language models \cite{kojima2022large}. However, the more important aspect of the reasoning is building trust of humans by providing additional information (the reasoning itself) that can be cross-checked and verified \cite{li2024llms, liang2024improving}. The development of interpretable models and methods for explaining their predictions is vital to building user trust and ensuring that these systems are used responsibly. Alongside explainability, fairness is a crucial component of trustworthy AI, requiring that these systems be free from biases that could lead to discriminatory outcomes, particularly in sensitive domains like healthcare \cite{huang2025survey}. Researchers are focusing on methods for detecting and mitigating biases, ensuring that LLMs treat all users equitably \cite{chaudhari2024rlhf, yang2024unmasking, cross2024bias}.

Moreover, accountability plays a central role in the trustworthiness of LLMs, particularly in high-stakes healthcare contexts where decisions can directly impact patient outcomes. LLMs are now being used more and more for different purposes, including therapy \cite{ren2023chatasd, bill2023fine}, diagnosis \cite{yang2024drhouse, kusa2023dr}, or assisting doctors \cite{bulut2024artificial, wu2023medical}. This requires clear guidelines for the responsible use of AI in medicine, including strong human oversight, clinical validation, and safeguards against misinformation or unsafe outputs. As LLMs become more integrated into diagnostic support, patient communication, and documentation, embedding ethical frameworks into their development and deployment will be essential. Such frameworks must align with core healthcare principles, such as beneficence, non-maleficence, autonomy, and justice, while addressing pressing concerns around data privacy, security, bias, and explainability.

Multimodal foundation models (MMFMs) are vital in areas like autonomous driving and healthcare but exhibit vulnerabilities such as biases and unsafe content \cite{xummdt}. The MMDT (Multimodal DecodingTrust) platform addresses these issues by comprehensively evaluating MMFMs' safety, fairness, privacy, and robustness, identifying areas for improvement \cite{xummdt}.

\section{Evaluation Methods and Metrics of Trustworthy AI}

\subsection{Evaluation methods}

Evaluating explainable AI methods is challenging due to the absence of ground truth in real-world data, unlike supervised learning tasks. Doshi-Velez and Kim (2017) \cite{doshi2017towards} propose three levels of evaluation: (1) Application level, where explanations are tested by end-users (e.g., radiologists using fracture detection software), (2) Human level, using simplified tasks with laypersons, and (3) Function level, relying on proxies such as model characteristics (e.g., tree depth). Evaluation further involves examining properties of explanation methods (e.g., expressiveness, translucency, portability, complexity) and individual explanations (e.g., accuracy, fidelity, stability, comprehensibility \cite{molnar2020interpretable}. These properties provide a framework for assessing how well explanations align with model predictions, user understanding, and task-specific requirements, emphasizing the importance of comprehensibility, certainty, and novelty in fostering trustworthy AI.

Z-Inspection is a method introduced by Zicari et al. (2021) \cite{zicari2021z} that integrates holistic and analytic approaches to evaluate the ethical impact of AI technologies on users, society, and the environment. Its process involves three phases: Set Up, which establishes the assessment framework; Assess, which identifies and maps ethical, technical, and legal issues; and Resolve, which addresses these issues and provides recommendations for ethical AI maintenance. Kowald et al. (2024) \cite{kowald2024establishing} provides a comprehensive perspective on trustworthy AI by focusing on evaluation aspects across the entire AI lifecycle. It offers a unified approach that addresses technical, human-centric, and legal requirements, including transparency, fairness, accountability, and human agency, while outlining open challenges and opportunities for further research.

\subsection{Metrics of Evaluation}

Evaluating trustworthy AI methods is essential to ensure their utility, reliability, and alignment with user needs. Effective metrics assess various aspects of explanations, such as their accuracy, simplicity, and robustness, providing a comprehensive understanding of the strengths and limitations of the methods. This subsection outlines key metrics, including validity, fidelity, consistency, sparsity, and robustness, which collectively capture the quality of explanations from different perspectives.

\subsubsection{Trust}
Schmidt and Biessman \cite{Schmidt2019} introduce a trust coefficient to quantify the extent to which human decisions are influenced by machine learning (ML) model predictions relative to ground truth labels:

\[
\bar{T} = \frac{\text{ITR}_{\hat{Y}_{\text{ML}}}}{\text{ITR}_Y}
\]

Here, $\text{ITR}_{\hat{Y}_{\text{ML}}}$ measures the mutual information between human decisions and model predictions, while $\text{ITR}_Y$ measures it with respect to the true labels. A coefficient greater than one suggests over-reliance on the model, whereas a value below one indicates skepticism or reduced trust. We refer the readers to the paper by Schmidt and Biessman \cite{Schmidt2019} for more details about the calculations of $\text{ITR}_{\hat{Y}_{\text{ML}}}$ and $\text{ITR}_Y$.

Survey-based assessments \cite{sisk2025validating} provide a validated framework for measuring public trust and openness toward AI in healthcare. They can evaluate attitudes across multiple dimensions, including diagnosis, treatment, and decision-making support, while also capturing concerns related to privacy, equity, care quality, and the human element of medicine. Such multi-dimensional measures offer critical insights into the factors that influence patients’ acceptance of AI-enabled technologies in learning health systems.

\subsubsection{Validity}

Validity refers to whether a generated counterfactual truly changes the model's prediction compared to the original input. Formally, a counterfactual example $\mathbf{x}'$ is considered valid if it results in a different predicted class than the original instance $\mathbf{x}$ \cite{guidotti2024counterfactual}, i.e.,
\[
b(\mathbf{x}') \neq b(\mathbf{x}),
\]
where $b(\cdot)$ denotes the prediction function.

According to Verma et al. \cite{verma2020counterfactual}, a valid counterfactual must be classified in the desired target class. To evaluate this quantitatively, Mothilal et al. \cite{mothilal2020explaining} define validity as the fraction of unique counterfactuals that produce a different outcome than the original input:
\[
\text{Validity} = \frac{|\{\mathbf{c} \in C \mid f(\mathbf{c}) \neq f(\mathbf{x})\}|}{k},
\]
where $C$ is the set of generated counterfactuals, $k$ is the total number generated, and $f(\cdot)$ is the model's prediction function. Only unique counterfactuals need to be considered to avoid overcounting identical examples returned by the method.

\subsubsection{Fidelity}

Fidelity measures how well an explanation approximates the black-box model’s predictions, and is crucial for trust, as low-fidelity explanations fail to reflect the model’s true behavior and thus lack utility \cite{molnar2020interpretable}. There are multiple computational measures for fidelity, such as simulated experiments, sanity checks, and comparative evaluation \cite{mohseni2021multidisciplinary}. Mir\'{o}-Nicolau et al. \cite{miro2025comprehensive} discussed four metrics of fidelity in their recent review paper: region perturbation, faithfulness correlation, faithfulness estimation, and infidelity. The LIME paper by Riberio et al. \cite{ribeiro2016should} provides a loss function for unfaithfulness of an explanation, which is the opposite of fidelity. An explanation is a simplified model \( g \in \mathcal{G} \) (e.g., a linear model or decision tree) that approximates the predictions of a complex model \( f \) in the vicinity of an instance \( x \), using interpretable components. To balance local fidelity and human interpretability, LIME minimizes the unfaithfulness of \( g \) to \( f \), while keeping the complexity \( \Omega(g) \) low:

\[
\xi(x) = \arg\min_{g \in \mathcal{G}} L(f, g, \pi_x) + \Omega(g)
\]

where \( L(f, g, \pi_x) \) is a loss function that measures the lack of fidelity between \( f \) and \( g \) in the locality defined by \( \pi_x \), and \( \Omega(g) \) is a measure of the complexity of the explanation model \cite{ribeiro2016should}.

\subsubsection{Proximity}

Mothilal et al. \cite{mothilal2020explaining} define proximity between a counterfactual example \( c_i \) and the original input \( x \) as the average feature-wise distance, computed separately for continuous and categorical features. For evaluation, they use the original feature scale (rather than the normalized scale used during CF generation) to enhance interpretability.

\[
\text{Proximity}_{\text{cont}} = -\frac{1}{k} \sum_{i=1}^{k} \text{dist}_{\text{cont}}(c_i, x)
\]

\[
\text{Proximity}_{\text{cat}} = 1 - \frac{1}{k} \sum_{i=1}^{k} \text{dist}_{\text{cat}}(c_i, x)
\]

In general, a higher proximity or a lower distance between the original example and its corresponding counterfactual example is preferred, so that the intervention to avoid an unwanted outcome (e.g., hyperglycemia) does not require drastic changes to the input features.

\subsubsection{Sparsity}

Sparsity measures how many features are changed in the counterfactual (CF) examples compared to the original input, providing a complementary perspective to proximity. Specifically, it quantifies the average fraction of unchanged features across all CF examples \cite{mothilal2020explaining}.

\[
\text{Sparsity} = 1 - \frac{1}{k \cdot d} \sum_{i=1}^{k} \sum_{l=1}^{d} 1_{[c_i^l \neq x^l]}
\]

In this equation, if the value of sparsity is high, the counterfactual will be more sparse, or fewer features will need to be changed. Because of that, the average number of mismatched features is subtracted from 1. The higher value of the sparsity metric according to this equation is better, because that would mean fewer feature changes and less burden on the user to adhere to the intervention.

While calculating the metric according to this equation help optimize counterfactuals, a more human-friendly metric often reported is how many features, on average, are changed \cite{mamun2024use}. It can be calculated by the following equation:

\[
\text{Features changed} = \frac{1}{k} \sum_{i=1}^{k} \sum_{l=1}^{d} 1_{[c_i^l \neq x^l]}
\]

Note that the division by the number of features is absent here, so that regardless of how many input features were used, the user can understand how many parameters would need to be changed.

\subsubsection{Diversity}
Different users of a system can have different preferences and abilities to apply interventions to overcome a health challenge. While counterfactuals need to be in close proximity to the original example, it is important for the counterfactuals of the same example to be diverse enough, when possible, so that the user can have different options.

Diversity of counterfactual (CF) examples is computed by measuring the average pairwise distance between CF examples, using either continuous or categorical feature-wise distance. A count-based diversity metric is also used that captures the average fraction of differing features between each pair of CFs \cite{mothilal2020explaining}.

\[
\text{Diversity} = \Delta = \frac{1}{\binom{k}{2}} \sum_{i=1}^{k-1} \sum_{j=i+1}^{k} \text{dist}(c_i, c_j)
\]

\[
\text{Count-Diversity} = \frac{1}{\binom{k}{2} \cdot d} \sum_{i=1}^{k-1} \sum_{j=i+1}^{k} \sum_{l=1}^{d} 1_{[c_i^l \ne c_j^l]}
\]

These equations take pairwise distances among the counterfactual examples and take the average by dividing the summation of the distances by the number of pairs, which is the number of choices for pairs from $k$ counterfactuals: calculated by $\binom{k}{2}$. For categorical values, the distance is calculated by adding the number of mismatched feature values and dividing by the number of features.

\section{Conclusion}

Building trustworthy AI systems for digital health requires addressing both robustness and explainability as foundational properties. This review synthesized recent developments across algorithmic innovations, evaluation frameworks, and practical deployment challenges to highlight the current landscape and emerging opportunities. Robustness in clinical settings must account for data distribution shifts, sensor noise, and adversarial scenarios, while explainability must go beyond model transparency to support clinician understanding, justification, and decision-making.

We emphasize that trustworthiness is not a monolithic concept, but a composite of verifiable properties that must be operationalized and measured systematically. As AI becomes increasingly integrated into healthcare workflows, aligning technical goals with human-centered requirements will be essential. Ongoing research in causal reasoning, uncertainty quantification, and interactive explanation offers promising paths forward. Ultimately, advancing trustworthy AI in digital health is not only a technical challenge but a societal imperative.

\bibliographystyle{plain}
\bibliography{root}

\end{document}